\documentclass[11pt]{article}
\usepackage{graphicx} 
\usepackage[a4paper,margin=1in]{geometry}
\usepackage{amsmath}
\usepackage{booktabs}
\usepackage{hyperref}
\usepackage{enumitem}
\usepackage{xcolor}
\usepackage{tabularx}
\usepackage{subcaption}
\usepackage{graphicx}
\usepackage{float}
\title{VDCook:DIY video data cook your MLLMs}
\author{Chengwei Wu, Hanyu Zhao, YiMing Ju, Li Du, Tengfei Pan \\ cwwu@baai.ac.cn}
\date{}

\begin{document}

\maketitle
\section{Abstract}
We introduce VDCook: a self-evolving video data operating system, a configurable video data construction platform for researchers and vertical domain teams. Users initiate data requests via natural language queries and adjustable parameters (scale, retrieval-synthesis ratio, quality threshold). The system automatically performs query optimization, concurrently running real video retrieval and controlled synthesis modules. It ultimately generates in-domain data packages with complete provenance and metadata, along with reproducible Notebooks.

Unlike traditional static, one-time-built datasets, VDCook enables continuous updates and domain expansion through its automated data ingestion mechanism based on MCP (Model Context Protocol)\cite{mcp2024anthropic}, transforming datasets into dynamically evolving open ecosystems. The system also provides multi-dimensional metadata annotation (scene segmentation, motion scoring, OCR ratio, automatic captioning, etc.), laying the foundation for flexible subsequent data `cooking' and indexing\cite{vlogger}.

This platform aims to significantly lower the barrier to constructing specialized video training datasets through infrastructure-level solutions, while supporting community contributions and a governance-enabled data expansion paradigm.
\textbf{Project demo:} \url{https://screenapp.io/app/v/WP0SvffgsH}

\section{Introduction}

Large-scale multimodal pretraining has fundamentally reshaped modern vision-language and video-language models\cite{emo}\cite{vexpress}. 
However, despite the abundance of publicly available video corpora\cite{miech2019howto100m}\cite{bain2021frozen}\cite{chen2024panda}\cite{wang2025koala}, constructing a high-quality \textit{in-domain} 
video dataset remains a costly and technically demanding process. 
Researchers and vertical teams often face several practical challenges: 
(i) existing datasets are either too large to locally process, 
(ii) domain-specific subsets are difficult to curate without rebuilding indexing pipelines, and 
(iii) data construction workflows are typically static and non-reproducible.

Traditional dataset construction follows a one-shot offline paradigm: 
data are crawled, cleaned, filtered, and packaged into a fixed release. 
Once published, the dataset becomes static, and adapting it to new domains 
requires repeating the entire pipeline. 
This paradigm does not scale well for rapidly evolving research demands 
or domain-specific applications.

In this work, we rethink video dataset construction as a \textbf{configurable and continuously evolving system}. 
We introduce \textbf{VDCook}, a self-evolving video data operating system that enables users to 
``cook'' their own in-domain video datasets on demand. 
Instead of distributing a fixed dataset, VDCook provides an infrastructure layer 
that allows users to specify a natural language query and configurable parameters 
(e.g., dataset scale, retrieval--synthesis ratio, quality thresholds), 
and automatically generates a tailored dataset through a modular processing pipeline.

The core idea of VDCook is to decouple \textit{data enrichment} from \textit{data filtering}. 
Each video clip is first processed through multiple metadata enrichment modules, 
including scene-based segmentation, motion scoring, OCR text ratio estimation, 
and automatic caption tagging. 
Rather than aggressively filtering data at preprocessing time, 
we retain these metadata attributes as flexible indexing signals, 
enabling downstream dataset cooking with minimal information loss. 
In practice, we only apply basic structural filtering (e.g., removing clips shorter than 2 seconds), 
while preserving rich annotations for later dynamic selection.

To further enable sustainable dataset growth, VDCook integrates an automated 
data ingestion mechanism based on a Modular Crawling Protocol (MCP), 
allowing new video sources to be continuously incorporated under standardized metadata specifications. 
This design transforms dataset construction from a static artifact into a living ecosystem, 
supporting dynamic updates, reproducibility, and community contribution.

The role of VDCook is mainly reflected in the following three aspects:

\begin{itemize}
    \item We propose a system-level rethinking of video dataset construction, 
    formulating it as an on-demand, configurable cooking process rather than a fixed offline release.
    
    \item We design a modular pipeline that combines query optimization, 
    parallel retrieval and controllable synthesis, and multi-dimensional metadata enrichment 
    to support flexible in-domain dataset generation.
    
    \item We introduce an extensible data ingestion and governance mechanism, 
    enabling continuous dataset evolution with provenance tracking and reproducibility.
\end{itemize}

By lowering the technical barrier to constructing domain-specific video datasets, 
VDCook aims to democratize multimodal data access and provide an infrastructure 
for scalable, reproducible, and continuously evolving video pretraining data.

\begin{figure}[t]
  \centering
  \includegraphics[width=1.1\linewidth]{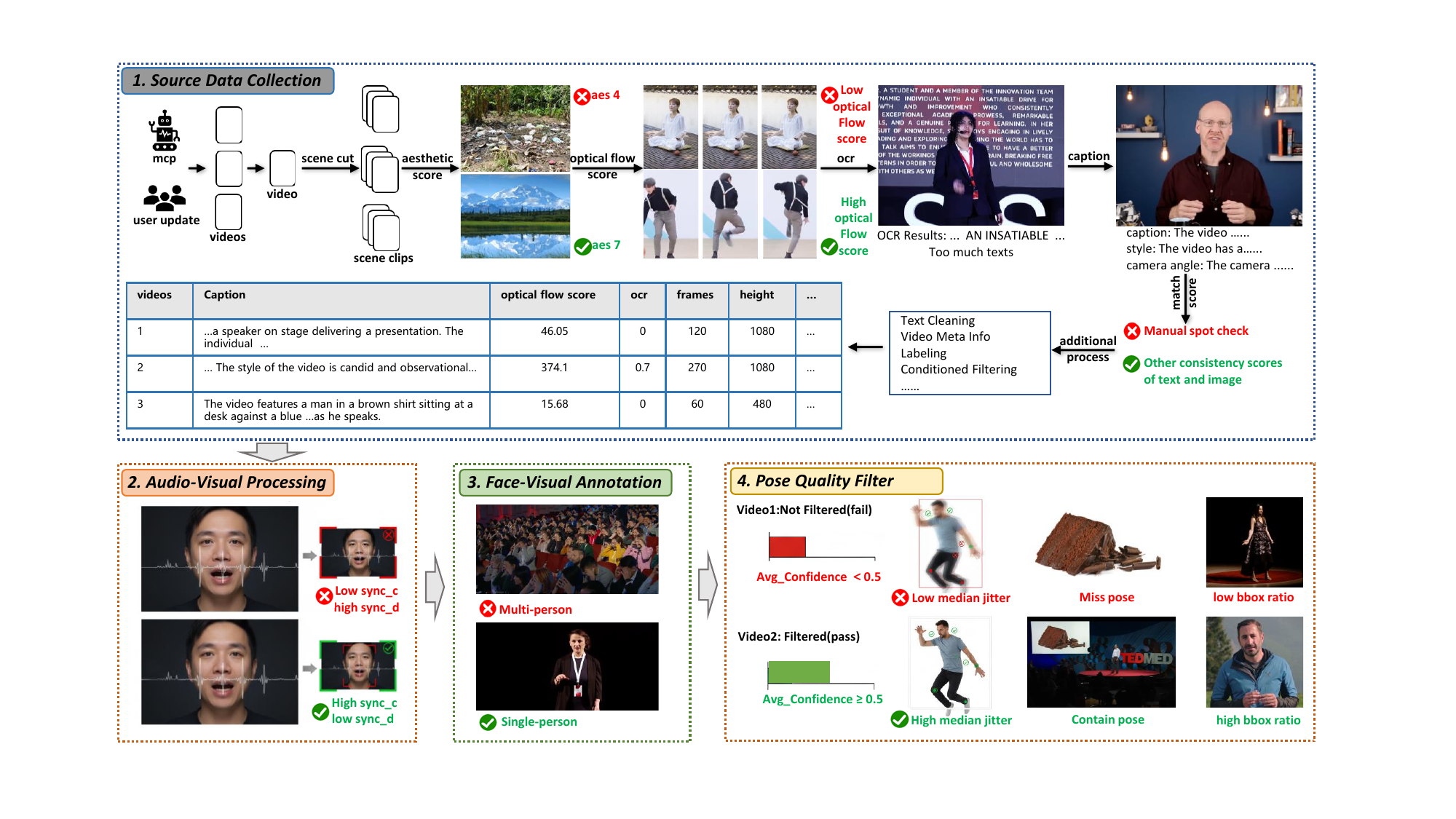}
  \caption{High-level architecture of VDCook. The system comprises automated ingestion (MCP), metadata enrichment modules, an index \& retrieval service, a controllable synthesis engine, quality \& provenance tracking, and a contribution/governance layer.}
  \label{fig:pipeline}
\end{figure}

\section{System Overview}

Figure~\ref{fig:pipeline} gives a high-level view of VDCook's architecture. \textbf{VDCook} is designed as a self-evolving data production engine rather than a static dataset repository. It integrates dynamic data acquisition, modular processing pipelines, large-scale model annotation, baseline evaluation, and long-tail data augmentation into a unified framework.

The system operates as a closed-loop architecture that continuously updates, refines, and expands in-domain video datasets according to user needs.

\subsection{Multi-Source Data Acquisition}

VDCook supports two primary data sources:

\textbf{(1) Automated Web Crawling via MCP.}
Through an MCP-based crawler, the system automatically retrieves videos from the web based on optimized user queries. The crawler supports:

\begin{itemize}
    \item Query optimization and expansion
    \item Domain-specific filtering
    \item Periodic re-crawling for dynamic updates
    \item Provenance tracking and metadata logging
\end{itemize}

This enables continuous data refresh and long-term evolution of domain datasets.

\textbf{(2) User-Contributed Data.}
Users may upload proprietary or private-domain video data. These data are processed through the same standardized pipeline, ensuring consistency across public and private sources.

Users can select which data sources to activate (MCP-crawled, user-uploaded, or hybrid mode), allowing flexible dataset construction strategies.

\subsection{Unified Video Processing Pipeline}

All videos, regardless of origin, are processed through a modular pipeline\cite{chen2017coherent}:

\begin{itemize}
    \item Scene segmentation\cite{castellano2020pyscenedetect} 
    \item Motion scoring and categorization\cite{lucas1981iterative}\cite{farneback2003two}    
    \item OCR text ratio estimation\cite{neudecker2021survey}
    \item Automatic video captioning
    \item Metadata standardization
\end{itemize}

Unlike traditional filtering-heavy pipelines, we preserve most processed data (except clips shorter than 2 seconds) and treat extracted signals as structured attributes rather than hard filters. This design enables downstream users to flexibly define selection criteria without permanently discarding data.

\subsection{Indexing, Retrieval and Cooking}
Annotated clips are ingested into a multi-tier index (metadata store + vector indices). The cooking stage performs the following:
\begin{enumerate}
  \item \textbf{Candidate retrieval:} the query-optimization agent expands user queries into multiple retrieval templates; retrieval pulls candidates across selected sources with prefilters (duration, language, safety flags).
  \item \textbf{Assembly and augmentation:} retrieved clips are assembled according to requested scale and retrieval--synthesis ratio; if synthesis is requested, the controllable synthesis engine uses selected seeds plus conditioning (keyframes, pose traces, style labels) to generate variants.
  \item \textbf{Policy filtering and packaging:} clips are filtered per the job's quality and compliance thresholds and packaged into a reproducible manifest with per-clip metadata (including MAC outputs and provenance).
\end{enumerate}

\subsection{Model Annotation Center}

A core component of VDCook is the \textbf{Model Annotation Center}, which integrates multiple large-scale models:

\begin{itemize}
    \item Vision-language models for captioning and semantic tagging
    \item Action recognition models
    \item OCR and scene text models
    \item Temporal consistency models
    \item Domain-specific fine-tuned models
\end{itemize}

This model ensemble allows multi-perspective labeling and structured tagging. Importantly, users can plug in custom models for domain adaptation.

The annotation center transforms raw videos into richly structured training-ready data.

\subsection{Baseline Evaluation Platform}

To bridge data quality and model performance, we integrate a baseline evaluation platform\cite{Activitynet}\cite{zhang2020inkthetics}
. 

Common benchmark models and evaluation metrics are provided for:

\begin{itemize}
    \item Video generation\cite{huang2024vbench} 
    \item Action recognition\cite{kay2017kinetics} 
    \item Multimodal dialogue\cite{maaz2023videochatgpt} 
    \item Temporal reasoning\cite{li2024mvbench} 
\end{itemize}

Users can train or fine-tune baseline models on the generated dataset and obtain quantitative feedback. The performance gap between baselines serves as an indirect signal of dataset effectiveness.

This mechanism connects data construction with downstream task performance.

\subsection{Long-Tail Data Augmentation via Strong Models}

A key innovation of VDCook is its long-tail data amplification mechanism.

For domains where real data is scarce (e.g., rare actions, niche artistic styles, specialized multimodal interactions), the system leverages strong generative models to synthesize new samples conditioned on:

\begin{itemize}
    \item Retrieved real clips
    \item Extracted structured attributes
    \item Domain prompts
    \item Style or action constraints
\end{itemize}

Generated data are treated as new data sources and re-enter the processing pipeline. They are annotated, evaluated, and benchmarked in the same way as real data.

This creates a \textbf{self-bootstrapping loop}:

\begin{center}
Real Data $\rightarrow$ Structured Processing $\rightarrow$ Model Training \\
$\rightarrow$ Synthetic Data Generation $\rightarrow$ Re-injection
\end{center}

Through this iterative mechanism, the system progressively improves coverage over long-tail distributions.

\subsection{Closed-Loop Data Evolution}

By combining dynamic crawling, user uploads, model annotation, baseline evaluation, and synthetic data injection, VDCook forms a data–model co-evolution loop.

The dataset is no longer static. It evolves with:

\begin{itemize}
    \item User queries
    \item Domain demands
    \item Model capability improvements
    \item Long-tail amplification
\end{itemize}

In this way, VDCook functions as an adaptive video data engine that continuously cooks, refines, and expands domain-specific datasets.

\section{Dataset Comparison and Statistics}

\begin{figure}[t]
  \centering
  \begin{subfigure}[b]{0.4\linewidth}
    \centering
    \includegraphics[width=\linewidth]{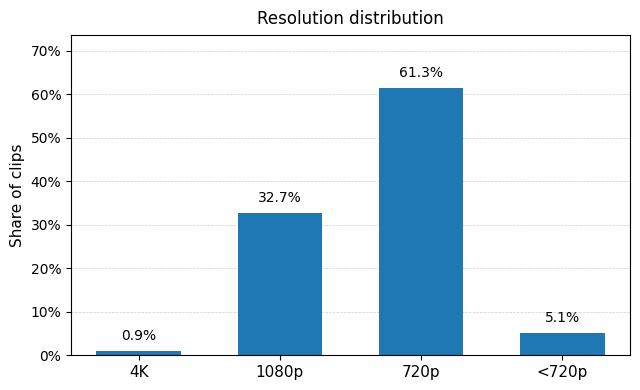}
    \caption{Resolution distribution.}
    \label{fig:resolution_dist}
  \end{subfigure}
  \hfill 
  \begin{subfigure}[b]{0.4\linewidth}
    \centering
    \includegraphics[width=\linewidth]{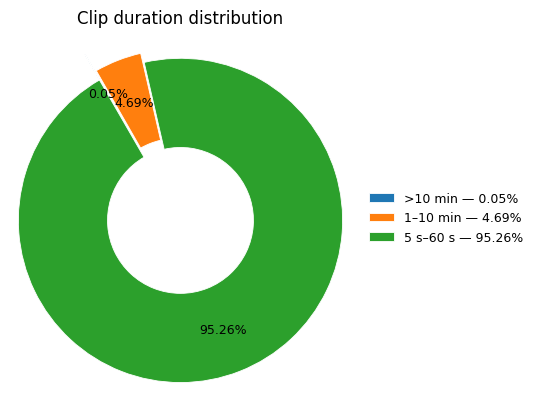}
    \caption{Duration distribution.}
    \label{fig:duration_dist}
  \end{subfigure}
  
  \caption{Analysis of our corpus: (a) Resolution distribution showing high-fidelity content; (b) Clip duration distribution primarily within 5--60\,s.}
  \label{fig:dataset_stats}
\end{figure}

To comprehensively characterize our 100M+ clip corpus and place it in context, Table~\ref{tab:dataset_overview} compares key summary statistics with several mainstream multimodal video datasets. Table~\ref{tab:dataset_meta} reports per-clip metadata statistics from our corpus (OCR coverage, motion intensity, OCR box counts). Figure~\ref{fig:resolution_dist} and Figure~\ref{fig:duration_dist} show the resolution and clip-duration distributions respectively.

\begin{table}[t]
  \centering
  \caption{High-level comparison between our corpus and representative video datasets. Numbers are reproduced from dataset metadata and internal summaries; please verify units (e.g., total duration) before final publication.}
  \label{tab:dataset_overview}
  \begin{tabular}{lcccc}
    \toprule
    Dataset & Overall scale & Avg. caption length (words) & Total duration & Typical resolution \\
    \midrule
    VidGen-1M\cite{tan2024vidgen} & 1M & 89.3 & --- & 720p \\
    MiraData\cite{ju2024miradata} & 330K & 318.0 & 16K & 720p \\
    Panda-70M\cite{chen2024panda} & 70M & 13.2 & 167K & 720p \\
    Koala-36M\cite{wang2025koala} & 36M & 202.1 & 172K & 720p \\
    \textbf{Ours} & \textbf{100M+} & \textbf{266} & \textbf{390K+} & 720p / 1080p \\
    \bottomrule
  \end{tabular}
  \vspace{2mm}
  \footnotesize{Notes: ``Total duration'' is reported in the original metadata (units as in source; verify whether hours/ minutes). ``Avg. caption length'' denotes mean caption length in words.}
\end{table}

\begin{table}[t]
  \centering
  \caption{Per-clip metadata statistics from our corpus (sampled/aggregated).}
  \label{tab:dataset_meta}
  \begin{tabularx}{\textwidth}{lcX} 
    \toprule
    Metric & value & Remarks \\
    \midrule
    ocr\_text\_area & avg:0.015, median:0.004 & Part of the data contains rich screen text, subtitles or title titles \\
    avg. motion intensity & 61.6 & A large number of videos with moderate to high-intensity exercise \\
    OCR boxes & 1.94 & High-density text information exists in some videos, which is valuable for training models that can handle complex visual-text interactions \\
    \bottomrule
  \end{tabularx}
  \vspace{1mm}
  \footnotesize{Notes: ``ocr\_text\_area'' is the fraction of frame area occupied by OCR bounding boxes averaged per clip. ``motion intensity'' is the platform's motion proxy score (see Appendix for exact formula).}
\end{table}

\paragraph{Interpretation and implications}
Several observations are noteworthy:
\begin{itemize}
  \item \textbf{Scale and caption richness.} Our corpus exceeds publicly available open datasets in raw clip count and provides long caption text on average (mean $\approx$ 266 words), which benefits instruction-style pretraining and dense captioning tasks.
  \item \textbf{High-fidelity visual inputs.} With a substantial fraction at 1080p (32.7\%) and some 4K content, the dataset supports high-resolution pretraining and tasks requiring fine-grained visual detail.
  \item \textbf{Text-heavy segments.} The mean OCR area (0.015) and average $\approx$1.94 OCR boxes per clip indicate non-trivial prevalence of screen text, subtitles, and title-frames—valuable for training robust visual–text interaction models.
  \item \textbf{Motion diversity.} The reported motion-intensity mean (61.6) and the duration distribution concentrated in 5--60\,s suggest the corpus contains many medium-length clips with substantial temporal dynamics, which is useful for temporal modeling, action recognition, and video-language alignment.
\end{itemize}

\paragraph{Recommended figures / analyses to include}
To make this subsection rigorous and convincing, include:
\begin{enumerate}
  \item A histogram / kernel density of caption lengths (words) and a table that reports percentiles (P10/P50/P90).
  \item Resolution bar chart (counts / fraction by resolution buckets: 360p/480p/720p/1080p/4K).
  \item Duration histogram (log-scale + percentiles).
  \item Distributions for OCR box counts and OCR area (violin / boxplot).
  \item Motion-intensity CDF and an example set of high/low-motion clip thumbnails.
  \item If possible, a small statistical comparison (e.g., KS test) between your corpus and the baselines for one or two key metrics to support claims of ``significant advantage.''
\end{enumerate}

\paragraph{Caveats and reproducibility}
Report the sampling procedure used to compute these aggregates (full-corpus vs random N-sample), the exact definitions (how motion intensity and OCR area are computed), and the snapshot date of ingestion. Include these details in the Appendix or caption so readers can reproduce the statistics.

\section{Cooked Datasets Across Domains}

To visually demonstrate the flexibility of VDCook, we present several representative domain-focused datasets constructed using our pipeline. Each case corresponds to a different data demand scenario and is illustrated with a $4 \times 3$ grid of sampled clips.

\subsection{Long-Tail Data}

Large-scale video corpora often under-represent rare but safety-critical or domain-specific events. 
Using VDCook, we construct multiple long-tail datasets covering urban risk scenarios and medical imaging content. 
Each category is retrieved through query expansion and curated through our unified pipeline. 
Representative samples are shown below.

\paragraph{Road Waterlogging}

\begin{figure}[H]
  \centering
  \includegraphics[width=\linewidth]{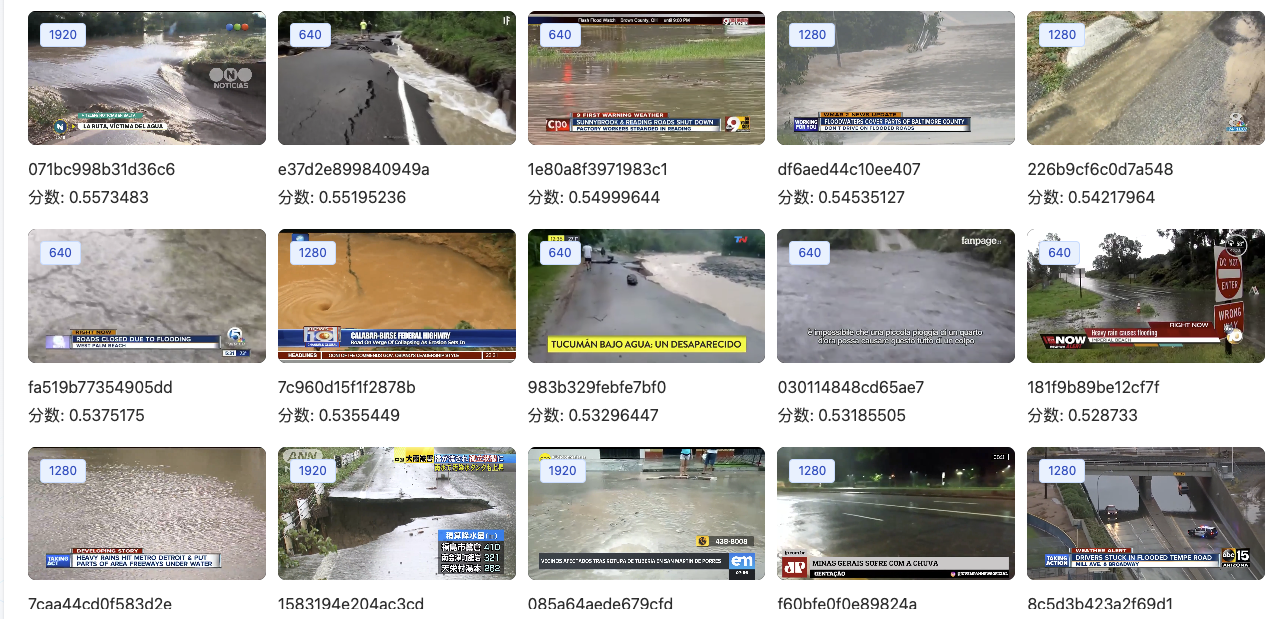}
  \caption{Representative samples of road waterlogging events. Such scenarios are rare in generic datasets but critical for urban risk monitoring and autonomous systems.}
  \label{fig:waterlogging}
\end{figure}

\paragraph{Dump Trucks in Construction Zones}

\begin{figure}[H]
  \centering
  \includegraphics[width=\linewidth]{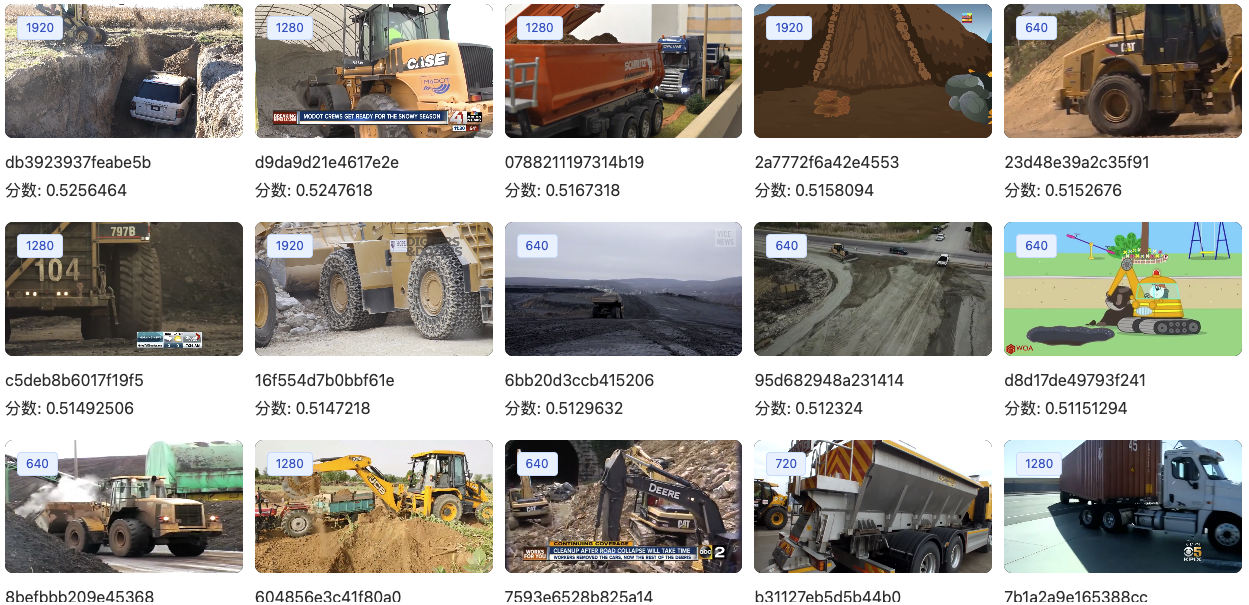}
  \caption{Examples of dump trucks in construction environments. These scenes involve heavy machinery and dynamic urban contexts, often underrepresented in general-purpose datasets.}
  \label{fig:dumptruck}
\end{figure}

\paragraph{Road Snow Accumulation}

\begin{figure}[H]
  \centering
  \includegraphics[width=\linewidth]{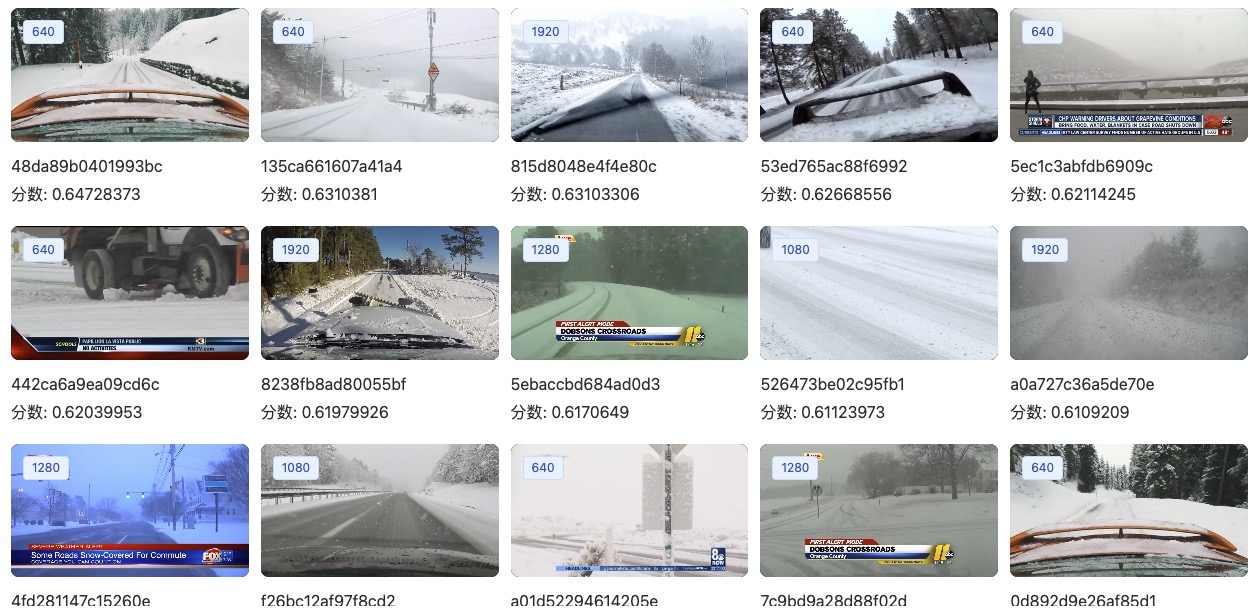}
  \caption{Clips of road snow accumulation under various lighting and weather conditions. Such data are important for transportation safety and seasonal robustness.}
  \label{fig:snow}
\end{figure}

\paragraph{Fallen Urban Greenery}

\begin{figure}[H]
  \centering
  \includegraphics[width=\linewidth]{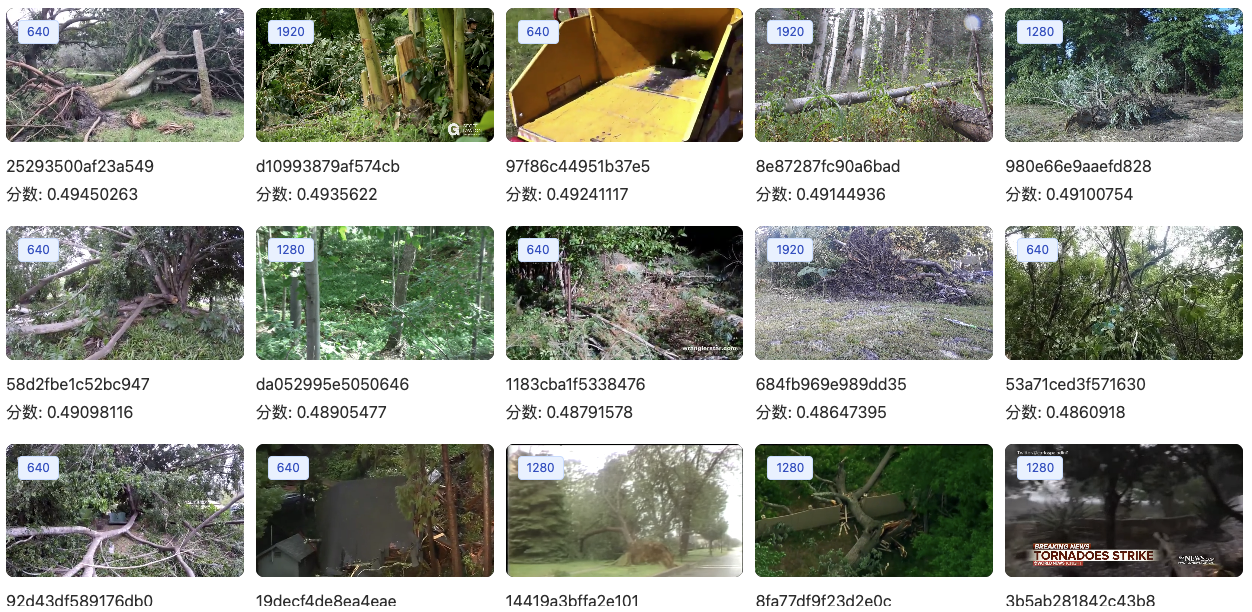}
  \caption{Examples of fallen trees or urban greenery after storms or accidents. These events are infrequent but important for disaster response modeling.}
  \label{fig:fallen_greenery}
\end{figure}

\paragraph{Pulmonary CT Angiography}

\begin{figure}[H]
  \centering
  \includegraphics[width=\linewidth]{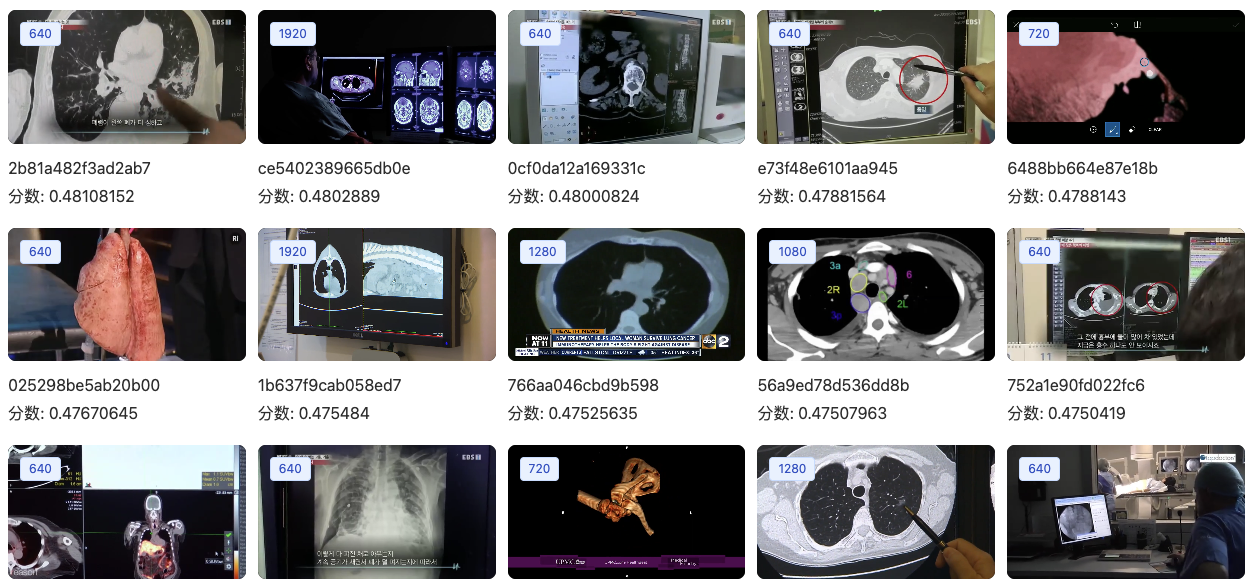}
  \caption{Representative pulmonary CT angiography video sequences. Medical imaging scenarios are highly domain-specific and scarce in open web video corpora.}
  \label{fig:ct}
\end{figure}

\subsection{Embodied Sequential Manipulation}

We construct embodied-action datasets that require multi-step reasoning and spatial grounding, such as:
``Pick up board 1 from the right side and place it on the rack; then pick up board 3 from the left side.''

\begin{figure}[H]
  \centering
  \includegraphics[width=\linewidth]{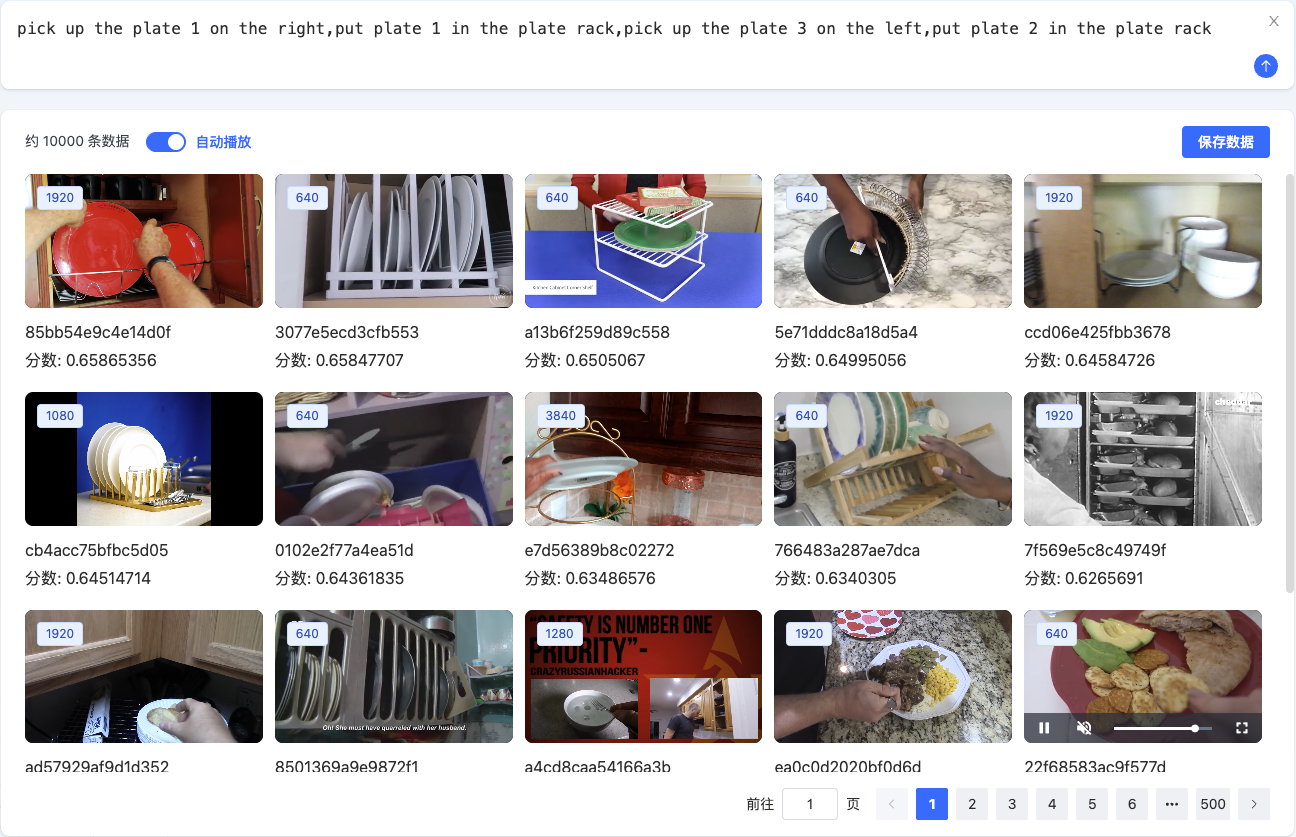}
  \caption{Examples of embodied sequential manipulation tasks. The clips highlight multi-step actions and spatial grounding.}
  \label{fig:embodied}
\end{figure}

The dataset emphasizes object tracking, left–right spatial understanding, and temporal consistency. Representative clips are shown in Figure~\ref{fig:embodied}.

\subsection{Multimodal Digital Humans}

This dataset focuses on face-centric videos with synchronized speech, gestures, and subtitles. It is designed for training conversational video models and multimodal dialogue systems\cite{celebv-text}\cite{styleheat}.

\begin{figure}[H]
  \centering
  \includegraphics[width=\linewidth]{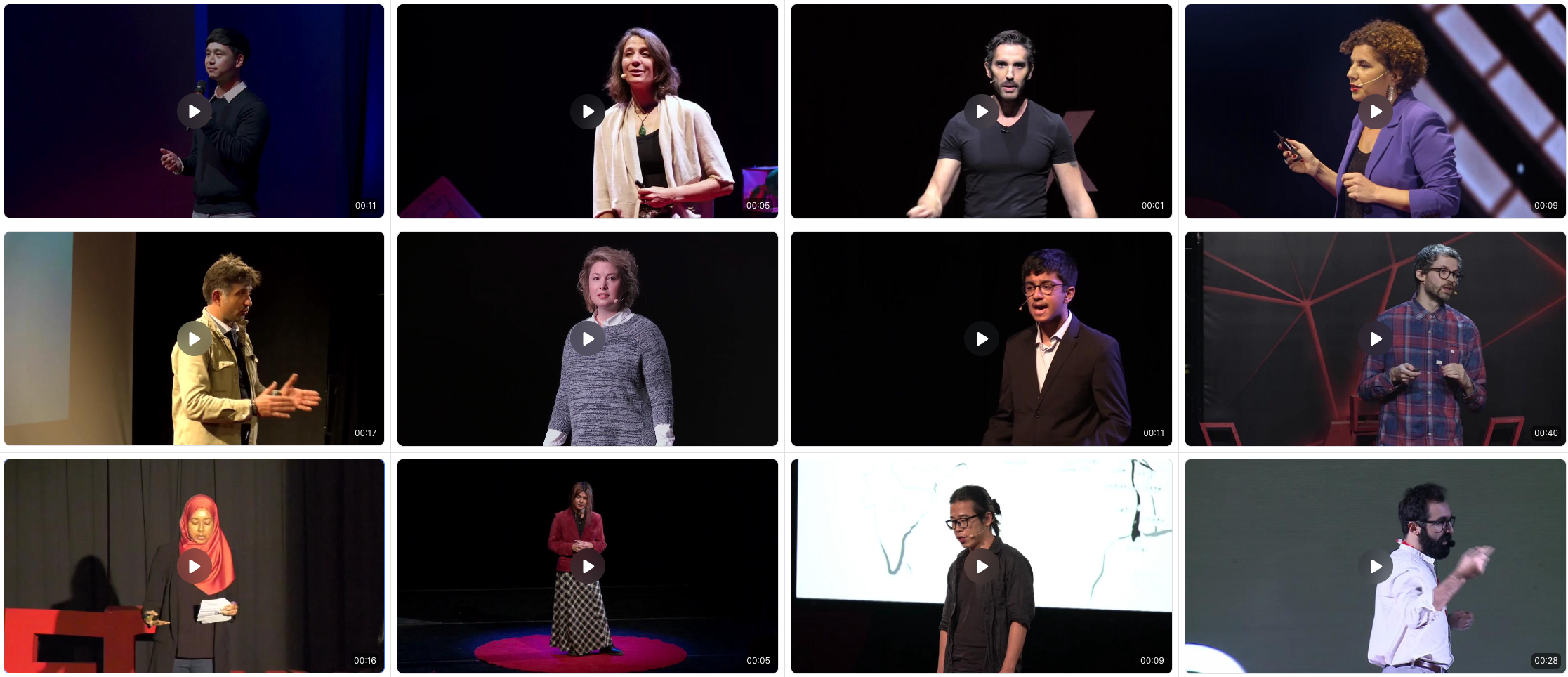}
  \caption{Multimodal digital human samples with speech, facial expression, and subtitle alignment.}
  \label{fig:digitalhuman}
\end{figure}

Figure~\ref{fig:digitalhuman} presents representative samples.

\subsection{Chinese Ink-Wash Style Video}

We curate and synthesize video clips in traditional Chinese ink-wash style. Due to the scarcity of real-world video materials in this artistic domain, synthetic augmentation plays an important role\cite{zhang2020inkthetics}.

\begin{figure}[H]
  \centering
  \includegraphics[width=\linewidth]{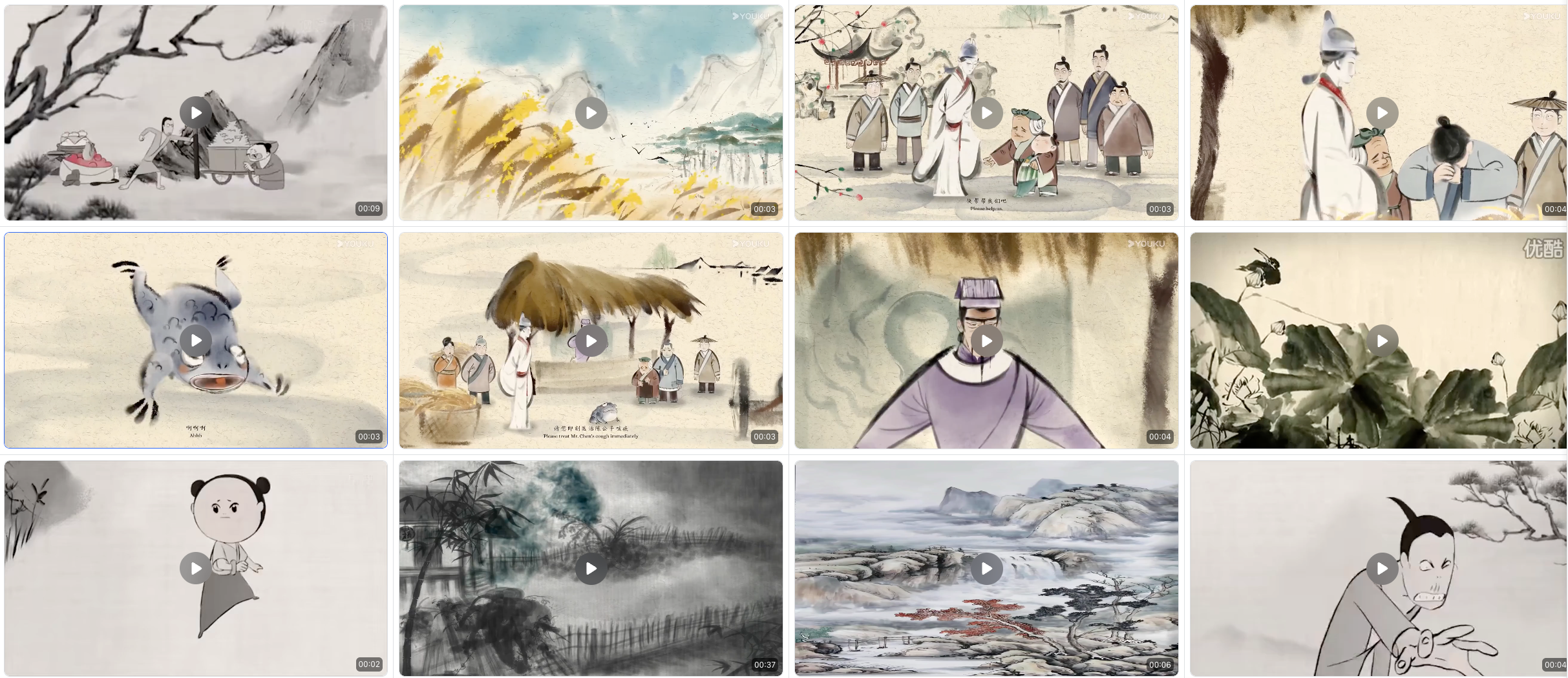}
  \caption{Chinese ink-wash style video samples. The dataset supports stylized generative modeling and temporal aesthetic learning.}
  \label{fig:inkwash}
\end{figure}

Examples are shown in Figure~\ref{fig:inkwash}.

\subsection{World Models and Physical Consistency}

We construct datasets emphasizing physically consistent interactions, including object collisions, gravity-driven motion, and fluid dynamics.

\begin{figure}[H]
  \centering
  \includegraphics[width=\linewidth]{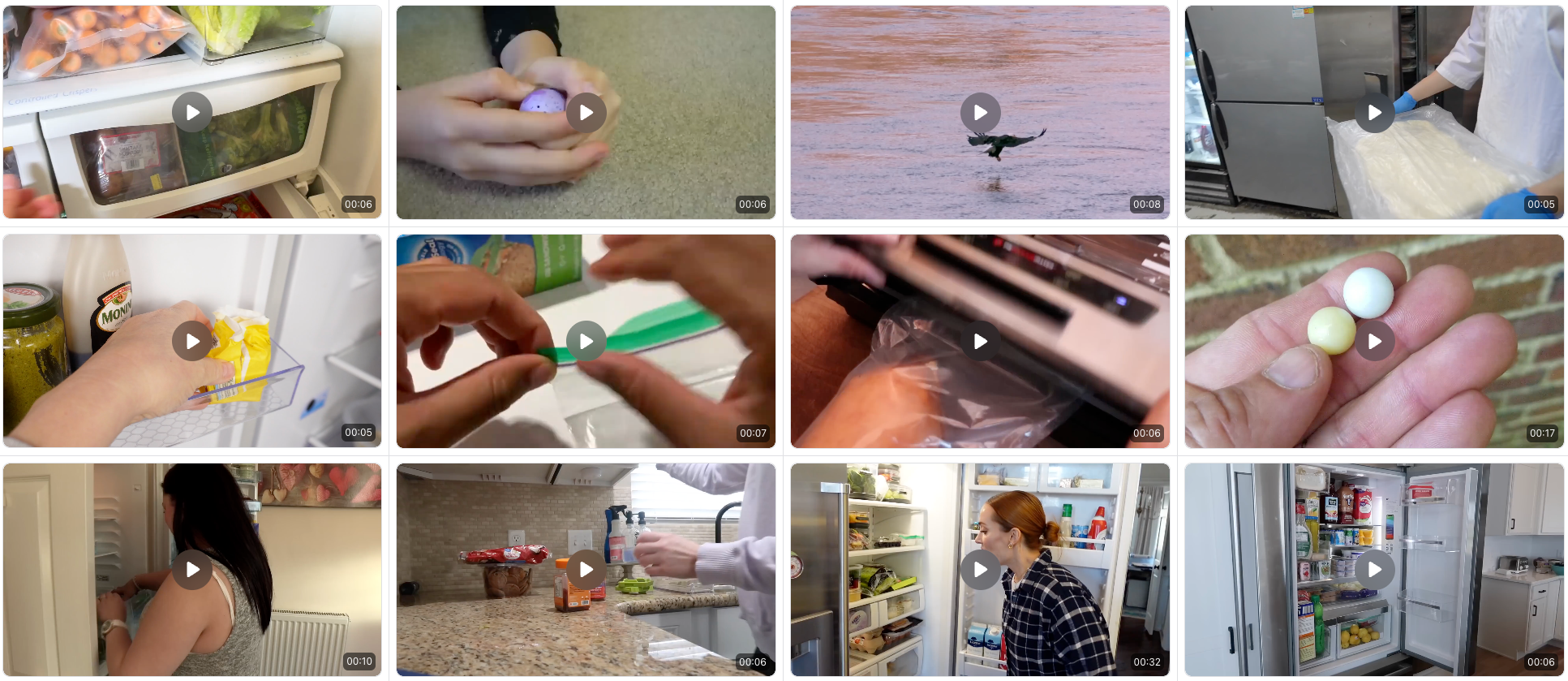}
  \caption{Examples of physics-driven interactions and temporally consistent object dynamics.}
  \label{fig:worldmodel}
\end{figure}

These clips are useful for training world models and evaluating physical reasoning capability. Representative samples are shown in Figure~\ref{fig:worldmodel}\cite{zhou2025negative}\cite{bar2024lumiere}\cite{shaulov2025flowmo}\cite{zheng2024open}.

\subsection{Model Validation: Ink-Wash Style Adaptation}

To validate the practical effectiveness of the constructed datasets, we conduct a style adaptation experiment using the Chinese ink-wash subset.

\paragraph{Experimental Setup}

We adopt Wan-1.3B\cite{wan2025wan} as the base generative model and fine-tune it on our curated ink-wash video subset. 
The goal is to evaluate whether domain-specific data constructed by VDCook can effectively guide stylistic adaptation toward traditional Chinese ink-wash aesthetics.

We compare outputs from the original base model and the fine-tuned model under identical text prompts.

\paragraph{Prompt 1: Mountain Landscape}

\begin{figure}[H]
  \centering
  \includegraphics[width=\linewidth]{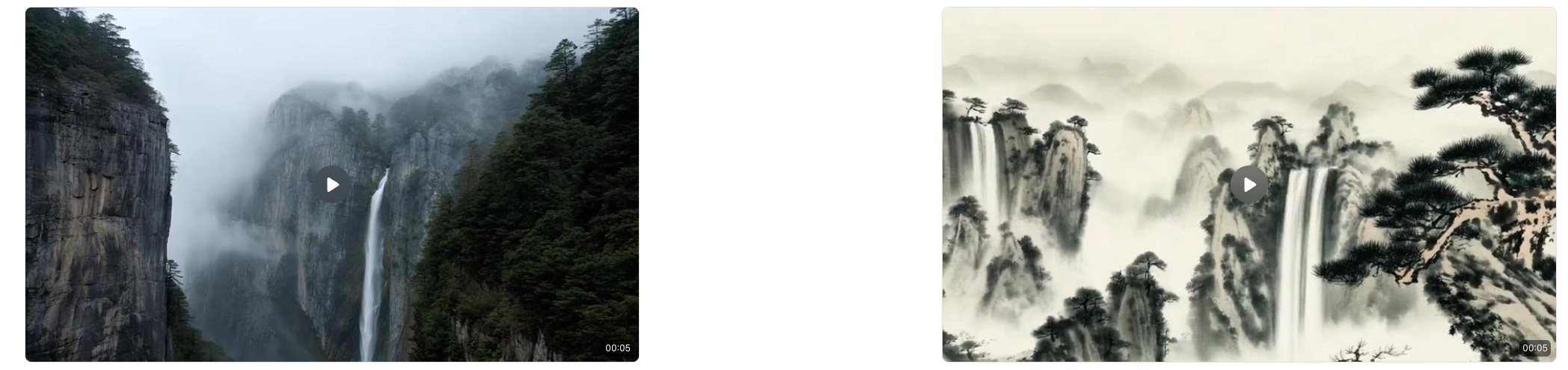}
  \caption{Left: Wan-1.3B base model output. 
  Right: Fine-tuned model on our ink-wash dataset.
  Prompt: ``A majestic mountain landscape with misty peaks, flowing waterfalls cascading down rocky cliffs, ancient pine trees swaying gently in the wind, birds flying across the vast sky, serene and peaceful atmosphere, traditional Chinese landscape painting composition, inkwash style.''}
  \label{fig:inkwash_compare1}
\end{figure}

\paragraph{Prompt 2: Plum Blossoms in Winter}

\begin{figure}[H]
  \centering
  \includegraphics[width=\linewidth]{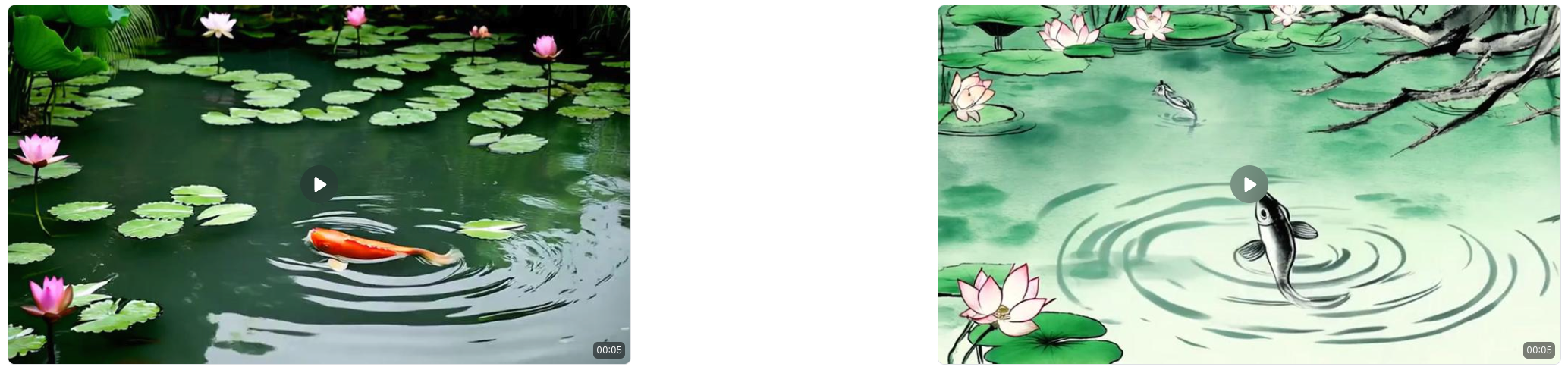}
  \caption{Left: Wan-1.3B base model output. 
  Right: Fine-tuned model.
  Prompt: ``Delicate plum blossoms blooming on ancient gnarled branches, petals gently falling in the cold winter wind, a small bird perched on a branch, snow-covered ground in the background, quiet winter scene with subtle beauty, inkwash style.''}
  \label{fig:inkwash_compare2}
\end{figure}

\paragraph{Prompt 3: Koi Fish in Garden Pond}

\begin{figure}[H]
  \centering
  \includegraphics[width=\linewidth]{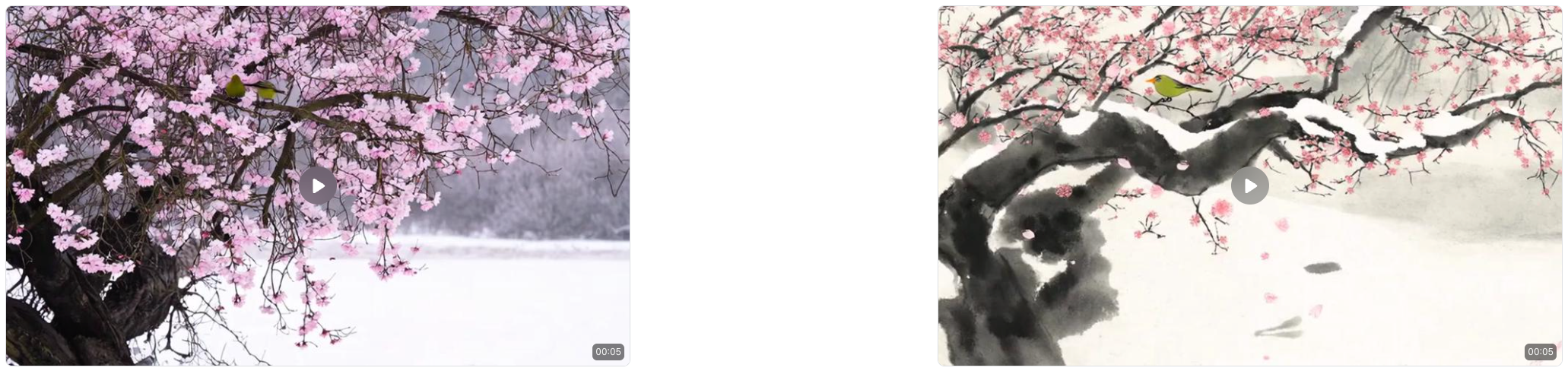}
  \caption{Left: Wan-1.3B base model output. 
  Right: Fine-tuned model.
  Prompt: ``Elegant koi fish swimming gracefully in a traditional Chinese garden pond, lotus flowers blooming on the water surface, ripples spreading across the calm water, weeping willow branches gently touching the water, serene garden atmosphere, inkwash style.''}
  \label{fig:inkwash_compare3}
\end{figure}

\paragraph{Observations}

Compared to the base model, the fine-tuned model demonstrates:

\begin{itemize}
    \item Stronger brushstroke-like texture representation
    \item Reduced photorealistic artifacts
    \item Improved ink diffusion effects
    \item More coherent traditional composition patterns
\end{itemize}

These results suggest that the curated ink-wash dataset effectively transfers domain-specific aesthetic priors into the generative model. 
The experiment provides qualitative evidence that the data constructed by VDCook is not only large-scale but also capable of guiding fine-grained stylistic adaptation.

\section{Conclusion}

We have presented VDCook (also referred to as VDCook), a self-evolving video data generation engine designed to lower the barrier for constructing in-domain video datasets. 
By combining modular ingestion (MCP + user uploads), multi-model metadata enrichment (Model Annotation Center), scalable indexing and retrieval, controllable synthesis for long-tail amplification, and an integrated baseline evaluation hub, the system enables interactive, reproducible, and domain-focused dataset “cooking” at large scale. 
Our 100M+ clip corpus and the curated sub-collections (e.g., ink-wash style, embodied manipulation, long-tail urban events) demonstrate the platform's ability to produce high-fidelity, task-relevant data; a small finetuning study on the ink-wash subset further validates the practical utility of the cooked data for stylistic adaptation.

We emphasize three practical design principles: (1) \emph{annotate-first, filter-later} to preserve flexible selection signals; (2) \emph{source-selectable ingestion} to balance freshness and fidelity; and (3) \emph{data–model co-evolution} where model-driven synthesis and baseline evaluation form a feedback loop that progressively improves coverage of rare phenomena. 
At the same time, we acknowledge current limitations: automated synthesis can introduce temporal artifacts, fine-grained provenance and licensing checks remain engineering-intensive, and rigorous quantitative benchmarks across diverse downstream tasks are an ongoing effort.

Looking forward, we plan to (i) formalize automated quality scoring and uncertainty estimation for per-clip selection, (ii) expand the Model Annotation Center with stronger domain adapters and calibrated confidence outputs, (iii) develop standardized quality- and ethics-aware export formats, and (iv) publish reproducible cook recipes, manifests, and demo notebooks to enable community validation and extension.

In sum, VDCook reframes dataset construction from a one-off release to a continuous, configurable service. We hope this work encourages a shift toward infrastructure-centered, reproducible data practices for multimodal research and application development.

\bibliography{main}

@misc{mcp2024anthropic,
  author       = {Anthropic},
  title        = {Model Context Protocol (MCP)},
  year         = {2024},
  howpublished = {\url{https://modelcontextprotocol.io}},
  note         = {Accessed: 2024-12-20}
}

@inproceedings{neudecker2021survey,
  title={A survey of OCR evaluation tools and metrics},
  author={Neudecker, Clemens and Baierer, Konstantin and Gerber, Mike and Clausner, Christian and Antonacopoulos, Apostolos and Pletschacher, Stefan},
  booktitle={Proceedings of the 6th International Workshop on Historical Document Imaging and Processing},
  pages={13--18},
  year={2021}
}

@article{castellano2020pyscenedetect,
  title={Pyscenedetect},
  author={Castellano, Brandon},
  journal={Last accessed},
  year={2020}
}

@inproceedings{miech2019howto100m,
  title={Howto100m: Learning a text-video embedding by watching hundred million narrated video clips},
  author={Miech, Antoine and Zhukov, Dimitri and Alayrac, Jean-Baptiste and Tapaswi, Makarand and Laptev, Ivan and Sivic, Josef},
  booktitle={Proceedings of the IEEE/CVF international conference on computer vision},
  pages={2630--2640},
  year={2019}
}

@inproceedings{bain2021frozen,
  title={Frozen in Time: A Joint Video and Image Encoder for End-to-End Retrieval},
  author={Bain, Max and Nagrani, Arsha and Varol, G{\"u}l and Zisserman, Andrew},
  booktitle={IEEE International Conference on Computer Vision (ICCV)},
  year={2021}
}

@inproceedings{chen2024panda,
  title={Panda-70m: Captioning 70m videos with multiple cross-modality teachers},
  author={Chen, Tsai-Shien and Siarohin, Aliaksandr and Menapace, Willi and Deyneka, Ekaterina and Chao, Hsiang-wei and Jeon, Byung Eun and Fang, Yuwei and Lee, Hsin-Ying and Ren, Jian and Yang, Ming-Hsuan and others},
  booktitle={Proceedings of the IEEE/CVF Conference on Computer Vision and Pattern Recognition},
  pages={13320--13331},
  year={2024}
}

@inproceedings{wang2025koala,
  title={Koala-36m: A large-scale video dataset improving consistency between fine-grained conditions and video content},
  author={Wang, Qiuheng and Shi, Yukai and Ou, Jiarong and Chen, Rui and Lin, Ke and Wang, Jiahao and Jiang, Boyuan and Yang, Haotian and Zheng, Mingwu and Tao, Xin and others},
  booktitle={Proceedings of the Computer Vision and Pattern Recognition Conference},
  pages={8428--8437},
  year={2025}
}

@inproceedings{lucas1981iterative,
  title={An iterative image registration technique with an application to stereo vision},
  author={Lucas, Bruce D and Kanade, Takeo},
  booktitle={Proceedings of the 7th International Joint Conference on Artificial Intelligence (IJCAI)},
  pages={674--679},
  year={1981}
}

@inproceedings{farneback2003two,
  title={Two-frame motion estimation based on polynomial expansion},
  author={Farneb{\"a}ck, Gunnar},
  booktitle={Scandinavian Conference on Image Analysis},
  pages={363--370},
  year={2003},
  organization={Springer}
}

@article{tan2024vidgen,
  title={Vidgen-1m: A large-scale dataset for text-to-video generation},
  author={Tan, Zhiyu and Yang, Xiaomeng and Qin, Luozheng and Li, Hao},
  journal={arXiv preprint arXiv:2408.02629},
  year={2024}
}

@article{ju2024miradata,
  title={Miradata: A large-scale video dataset with long durations and structured captions},
  author={Ju, Xuan and Gao, Yiming and Zhang, Zhaoyang and Yuan, Ziyang and Wang, Xintao and Zeng, Ailing and Xiong, Yu and Xu, Qiang and Shan, Ying},
  journal={Advances in Neural Information Processing Systems},
  volume={37},
  pages={48955--48970},
  year={2024}
}

@inproceedings{huang2024vbench,
  title={VBench: Comprehensive Benchmark Suite for Video Generative Models},
  author={Huang, Zanyi and others},
  booktitle={CVPR},
  year={2024}
}

@article{kay2017kinetics,
  title={The kinetics human action video dataset},
  author={Kay, Will and others},
  journal={arXiv preprint arXiv:1705.06950},
  year={2017}
}

@inproceedings{maaz2023videochatgpt,
  title={Video-ChatGPT: Towards Detailed Video Understanding via Large Language Models},
  author={Maaz, Muhammad and others},
  booktitle={ACL},
  year={2024}
}

@inproceedings{li2024mvbench,
  title={MVBench: A Comprehensive Multi-modal Video Understanding Benchmark},
  author={Li, Kunchang and others},
  booktitle={CVPR},
  year={2024}
}

@inproceedings{celebv-text,
  title={Celebv-text: A large-scale facial text-video dataset},
  author={Yu, Jianhui and Zhu, Hao and Jiang, Liming and Loy, Chen Change and Cai, Weidong and Wu, Wayne},
  booktitle={Proceedings of the IEEE/CVF Conference on Computer Vision and Pattern Recognition},
  pages={14805--14814},
  year={2023}
}

@inproceedings{vlogger,
  title={Vlogger: Make your dream a vlog},
  author={Zhuang, Shaobin and Li, Kunchang and Chen, Xinyuan and Wang, Yaohui and Liu, Ziwei and Qiao, Yu and Wang, Yali},
  booktitle={Proceedings of the IEEE/CVF Conference on Computer Vision and Pattern Recognition},
  pages={8806--8817},
  year={2024}
}

@inproceedings{emo,
  title={Emo: Emote portrait alive generating expressive portrait videos with audio2video diffusion model under weak conditions},
  author={Tian, Linrui and Wang, Qi and Zhang, Bang and Bo, Liefeng},
  booktitle={European Conference on Computer Vision},
  pages={244--260},
  year={2024},
  organization={Springer}
}

@article{vexpress,
  title={V-express: Conditional dropout for progressive training of portrait video generation},
  author={Wang, Cong and Tian, Kuan and Zhang, Jun and Guan, Yonghang and Luo, Feng and Shen, Fei and Jiang, Zhiwei and Gu, Qing and Han, Xiao and Yang, Wei},
  journal={arXiv preprint arXiv:2406.02511},
  year={2024}
}

@inproceedings{styleheat,
  title={Styleheat: One-shot high-resolution editable talking face generation via pre-trained stylegan},
  author={Yin, Fei and Zhang, Yong and Cun, Xiaodong and Cao, Mingdeng and Fan, Yanbo and Wang, Xuan and Bai, Qingyan and Wu, Baoyuan and Wang, Jue and Yang, Yujiu},
  booktitle={European conference on computer vision},
  pages={85--101},
  year={2022},
  organization={Springer}
}

@inproceedings{activitynet,
  title={Activitynet: A large-scale video benchmark for human activity understanding},
  author={Caba Heilbron, Fabian and Escorcia, Victor and Ghanem, Bernard and Carlos Niebles, Juan},
  booktitle={Proceedings of the ieee conference on computer vision and pattern recognition},
  pages={961--970},
  year={2015}
}

@article{zhang2020inkthetics,
  title={Inkthetics: a comprehensive computational model for aesthetic evaluation of Chinese ink paintings},
  author={Zhang, Jiajing and Miao, Yongwei and Zhang, Junsong and Yu, Jinhui},
  journal={IEEE Access},
  volume={8},
  pages={225857--225871},
  year={2020},
  publisher={IEEE}
}

@article{zhou2025negative,
  title={Negative Shanshui: Real-time Interactive Ink Painting Synthesis},
  author={Zhou, Aven-Le},
  journal={arXiv preprint arXiv:2508.16612},
  year={2025}
}

@inproceedings{bar2024lumiere,
  title={Lumiere: A space-time diffusion model for video generation},
  author={Bar-Tal, Omer and Chefer, Hila and Tov, Omer and Herrmann, Charles and Paiss, Roni and Zada, Shiran and Ephrat, Ariel and Hur, Junhwa and Liu, Guanghui and Raj, Amit and others},
  booktitle={SIGGRAPH Asia 2024 Conference Papers},
  pages={1--11},
  year={2024}
}

@article{shaulov2025flowmo,
  title={FlowMo: Variance-Based Flow Guidance for Coherent Motion in Video Generation},
  author={Shaulov, Ariel and Hazan, Itay and Wolf, Lior and Chefer, Hila},
  journal={arXiv preprint arXiv:2506.01144},
  year={2025}
}

@article{wan2025wan,
  title={Wan: Open and advanced large-scale video generative models},
  author={Wan, Team and Wang, Ang and Ai, Baole and Wen, Bin and Mao, Chaojie and Xie, Chen-Wei and Chen, Di and Yu, Feiwu and Zhao, Haiming and Yang, Jianxiao and others},
  journal={arXiv preprint arXiv:2503.20314},
  year={2025}
}

@article{zheng2024open,
  title={Open-sora: Democratizing efficient video production for all},
  author={Zheng, Zangwei and Peng, Xiangyu and Yang, Tianji and Shen, Chenhui and Li, Shenggui and Liu, Hongxin and Zhou, Yukun and Li, Tianyi and You, Yang},
  journal={arXiv preprint arXiv:2412.20404},
  year={2024}
}

@inproceedings{chen2017coherent,
  title={Coherent online video style transfer},
  author={Chen, Dongdong and Liao, Jing and Yuan, Lu and Yu, Nenghai and Hua, Gang},
  booktitle={Proceedings of the IEEE international conference on computer vision},
  pages={1105--1114},
  year={2017}
}

\end{document}